\documentclass[conference]{IEEEtran}
\usepackage{amsmath}

\IEEEoverridecommandlockouts

\usepackage{cite}
\usepackage{graphicx}
\usepackage{subcaption}
\usepackage{amsmath,amssymb}
\usepackage{booktabs}
\usepackage{float}
\usepackage{hyperref}
\usepackage{array}

\title{Benchmarking Machine Learning Models for Multi-Omics-Based Breast Cancer Prediction}

\author{
\IEEEauthorblockN{Priyanka Paudel and Madan Baduwal}
\IEEEauthorblockA{\textit{Department of Computer Science and Engineering, Mississippi State University,}
Mississippi State, MS, USA \\
pp918@msstate.edu, mb4239@msstate.edu}
}

\begin{document}

\maketitle

\begin{abstract}
Estrogen Receptor (ER) status is a critical biomarker in breast cancer diagnosis, prognosis, and treatment selection. Recent advances in high-throughput sequencing technologies have enabled the generation of multi-omics datasets that provide complementary molecular information for computational prediction tasks. This study presents a systematic benchmarking analysis of classical machine learning models for ER status prediction using transcriptomic (RNA expression), genomic (copy number variation; CNV), and proteomic (RPPA) data from the TCGA-BRCA cohort. A rigorous experimental framework incorporating stratified train-test splitting, stratified five-fold cross-validation, class imbalance handling, and fold-specific feature selection was employed to ensure reliable evaluation and prevent data leakage. Random Forest, XGBoost, LightGBM, CatBoost, Support Vector Machines (SVM), and Logistic Regression were evaluated across single-omic and multi-omic settings. Results demonstrated that RNA expression provided the strongest predictive signal, while multi-omic integration yielded modest but consistent improvements over individual modalities. Among all evaluated approaches, Random Forest achieved the best overall performance in the integrated multi-omic setting, obtaining a balanced accuracy of 90.3\% and an ROC-AUC of 97.1\%. Furthermore, recurrent selection of biologically relevant genes, including \textit{ESR1}, \textit{PGR}, \textit{FOXA1}, and \textit{GATA3}, supported the biological validity of the learned models. These findings indicate that carefully regularized classical machine learning methods remain highly effective for small, high-dimensional genomic datasets and that multi-omic integration provides complementary information for breast cancer ER status prediction.

\end{abstract}


\noindent\textbf{Code Availability:}
\url{https://github.com/priyanka36/MultiOmics-ER-Prediction}

\section{Introduction}

Breast cancer remains one of the most prevalent and clinically heterogeneous cancers worldwide. Molecular heterogeneity within breast tumors leads to substantial differences in prognosis, treatment response, and survival outcomes among patients. Biomarkers such as Estrogen Receptor (ER), Progesterone Receptor (PR), and HER2 status are therefore essential components of modern breast cancer diagnosis and therapeutic decision-making.

Among these biomarkers, ER status is particularly important because it directly influences endocrine therapy selection and clinical management strategies. ER-positive and ER-negative tumors exhibit substantially different biological behaviors and transcriptional programs. Consequently, computational approaches capable of accurately predicting ER status from molecular data may contribute to improved precision medicine workflows and deeper biological understanding of breast cancer subtypes.

Recent advances in high-throughput sequencing technologies have enabled the large-scale generation of multi-omics datasets containing genomic, transcriptomic, proteomic, and clinical information~\cite{tcga2012,yoon2020}. Public repositories such as The Cancer Genome Atlas (TCGA) provide comprehensive molecular profiles of cancer patients and have become important resources for computational cancer research. However, effectively modeling multi-omics datasets presents several computational challenges, including high dimensionality, feature redundancy, class imbalance, and limited sample size.

Machine learning methods have demonstrated strong performance on high-dimensional biomedical datasets ~\cite{mogcn2022} due to their ability to capture nonlinear biological relationships. Ensemble tree methods and kernel-based models in particular have shown strong applicability to genomic prediction tasks. Although deep learning approaches have become dominant in many artificial intelligence domains, their application to small genomic datasets remains challenging because deep architectures typically require substantially larger sample sizes for reliable generalization.

This study therefore focuses primarily on systematically benchmarking classical machine learning models for ER status prediction using multiple omics modalities from TCGA-BRCA. The work investigates both single-omic and multi-omic learning strategies while emphasizing robust experimental methodology, biological interpretability, and reliable evaluation practices appropriate for small genomic datasets

\section{Research Objectives}

This study aims to systematically evaluate the effectiveness of machine learning approaches for predicting Estrogen Receptor (ER) status in breast cancer using multi-omics data from the TCGA-BRCA cohort. Given the heterogeneous nature of breast cancer and the availability of complementary molecular information from transcriptomic, genomic, and proteomic sources, the work investigates how different omics modalities contribute to predictive performance both individually and when integrated into a unified framework.

To achieve this goal, RNA expression, copy number variation (CNV), and protein expression (RPPA) data are analyzed separately to assess their individual predictive value, followed by a multi-omic integration strategy designed to determine whether combining modalities provides additional information beyond single-omic models. The study further compares several widely used machine learning algorithms, including Random Forest, XGBoost, LightGBM, CatBoost, Support Vector Machines, and Logistic Regression, to identify models that offer strong predictive performance while maintaining robustness and interpretability.

In addition to model performance, the study examines the biological significance and consistency of features selected across cross-validation folds. Particular attention is given to ensuring a rigorous evaluation framework through stratified train-test splitting, stratified cross-validation, fold-specific feature selection, class imbalance handling, and the use of multiple complementary performance metrics. Through this analysis, the study seeks to determine whether multi-omic integration improves ER status prediction and to identify a modeling strategy that balances predictive accuracy, generalization ability, and biological interpretability.

\section{Dataset Description}

This study utilized publicly available breast cancer multi-omics data obtained from the TCGA-BRCA cohort through the UCSC Xena Browser. The dataset provides matched molecular profiles ~\cite{tcga2012} for breast cancer patients across multiple biological layers, enabling the investigation of complementary genomic, transcriptomic, and proteomic information within a unified analytical framework. Three omics modalities were included in this study: RNA expression data representing transcriptomic activity, copy number variation (CNV) data reflecting genomic alterations, and protein expression measurements obtained through Reverse Phase Protein Array (RPPA) technology. 

The original cohort contained 705 patient samples. Following data cleaning and the removal of samples with ambiguous or unavailable ER-status annotations, a total of 549 patients remained for downstream analysis. The resulting dataset consisted of 604 RNA expression features, 860 CNV features, and 223 protein expression features, yielding a combined multi-omic feature space of 1,687 molecular variables. The prediction task was formulated as a binary classification problem, where each patient was categorized as either ER-positive or ER-negative. 

The final cohort exhibited a moderate class imbalance, with approximately 75.4\% of samples belonging to the ER-positive class and 24.6\% belonging to the ER-negative class. Because imbalanced class distributions can bias conventional performance measures, this study employed balanced evaluation metrics and class-weighted learning strategies to ensure reliable assessment of model performance across both classes. 

\section{Challenges of Multi-Omics Data}

Several important computational and biological challenges are associated with multi-omics genomic datasets.

\noindent$\bullet$\textbf{High Dimensionality:} The combined dataset contained 1,687 molecular features but only 549 usable patient samples. This creates a high-dimensional small-sample learning problem that substantially increases the risk of overfitting. Many machine learning models may memorize noise rather than learning biologically meaningful patterns under these conditions.

\noindent$\bullet$\textbf{Class Imbalance:} The ER-positive class substantially outnumbered the ER-negative class. Standard accuracy alone therefore becomes misleading because a naive classifier predicting all samples as ER-positive could still achieve high apparent accuracy without clinically meaningful discrimination.

\noindent$\bullet$\textbf{Small Sample Size:} Although TCGA provides biologically rich data, the effective number of usable samples remains relatively limited for complex machine learning models, especially deep learning architectures that typically require much larger datasets.

\noindent$\bullet$\textbf{Feature Redundancy and Correlation:} Genes participating in shared biological pathways are often highly correlated. Such redundancy can increase noise and reduce model generalization if not handled appropriately through feature selection and regularization.

\noindent$\bullet$\textbf{Data Leakage Risk:} Improper preprocessing can accidentally allow information from test data to influence training procedures, leading to overly optimistic evaluation metrics. Preventing leakage is especially critical in high-dimensional genomic datasets.

\section{Data Preprocessing Pipeline}

A structured preprocessing pipeline was developed to ensure reliable model training and unbiased performance evaluation. Particular attention was given to minimizing data leakage, handling class imbalance, and maintaining consistency across all experiments involving RNA expression, CNV, protein expression, and multi-omic datasets.

\paragraph{\textbf{Label Cleaning}}

The original TCGA-BRCA cohort contained 705 patient samples. To ensure reliable supervision during model training, only samples with clearly defined ER-status annotations were retained. Samples labeled as NaN, Indeterminate, Not Performed, or Performed but Not Available were excluded because such labels introduce ambiguity and may negatively affect model learning. Following this cleaning process, 549 patients remained for downstream analysis.

\paragraph{\textbf{Label Encoding}}

To facilitate binary classification, ER-status labels were converted into numerical form, with ER-negative samples encoded as 0 and ER-positive samples encoded as 1. This representation ensured compatibility across all machine learning algorithms while preserving the clinical interpretation of the target variable.

\paragraph{\textbf{Stratified Train-Test Split}}

The cleaned dataset was divided into training and testing subsets using an 80/20 stratified split. Stratification was employed to preserve the original distribution of ER-positive and ER-negative samples in both subsets, thereby reducing the risk of biased performance estimates that can arise from uneven class allocation. This approach ensured that the held-out test set remained representative of the overall cohort.

\paragraph{\textbf{Feature Selection}}

Given the high-dimensional nature of the dataset, feature selection was performed using ANOVA F-statistics implemented through \texttt{SelectKBest}. To prevent information leakage, feature selection was applied independently within each training fold during cross-validation rather than before data splitting. For RNA, CNV, RPPA, and multi-omic experiments, the top 300 features were selected within each fold, providing a balance between dimensionality reduction and retention of biologically relevant information.

\paragraph{\textbf{Feature Scaling}}

Feature scaling was applied only when required by the underlying learning algorithm. Specifically, \texttt{StandardScaler} normalization was used for Support Vector Machines and Logistic Regression because these methods are sensitive to differences in feature magnitude. In contrast, tree-based ensemble models such as Random Forest, XGBoost, LightGBM, and CatBoost were trained using the original feature values, as decision-tree-based methods are inherently scale-invariant.

\section{Experimental Strategy}

A progressive experimental strategy was adopted in order to systematically evaluate the predictive contribution of each omics modality before performing multi-omic integration.

Rather than directly combining all modalities at the beginning of the study, each omics layer was first evaluated independently. This design enabled clearer interpretation of predictive contributions and avoided masking weaker modalities within the combined feature space. The study was therefore divided into four sequential experimental phases.

\noindent\textbf{RNA-Only Experiments.} \hspace{0.2cm} RNA expression data was evaluated first because ER signaling is fundamentally transcription-driven. Since ER functions as a transcription factor, gene expression profiles were expected to contain the strongest direct representation of ER activity. The RNA-Only experiments therefore served as the primary baseline for the study.

\noindent\textbf{Copy Number Variation (CNV) Experiments.} \hspace{0.2cm} The second phase investigated copy number variation features independently. CNV data reflects structural genomic alterations rather than direct transcriptional activity. Consequently, CNV was expected to provide weaker but potentially complementary predictive information.

\noindent\textbf{Protein Expression Experiments.} \hspace{0.2cm} Protein expression data was then evaluated independently using RPPA-derived features. Protein-level measurements reflect downstream pathway activity and may capture biological states not fully represented at the transcriptomic level.

\noindent\textbf{Multi-Omic Integration.} \hspace{0.2cm}Finally, RNA, CNV, and protein features were combined into a unified feature space for multi-omic learning experiments. 

The primary objective of this phase was to determine whether integrating multiple molecular modalities improves ER status prediction beyond single-omic baselines.

\section{Why Classical Machine Learning Before Deep Learning}

Although deep learning approaches have demonstrated strong success in domains such as computer vision and natural language processing, genomic datasets present substantially different constraints. The dataset used in this study contained only 549 usable patient samples. Deep neural networks typically require substantially larger datasets to generalize reliably. Under small-sample high-dimensional settings, deep architectures are highly susceptible to overfitting and unstable optimization. In contrast, classical machine learning methods provide several advantages for genomic tabular data:

\begin{itemize}
    \item Ensemble tree methods naturally regularize through bagging and feature subsampling.
    \item SVMs effectively model nonlinear relationships using kernel functions.
    \item Classical models are computationally efficient on small datasets.
    \item Feature importance analysis improves biological interpretability.
    \item Conservative hyperparameter control is easier to implement.
\end{itemize}

These advantages make classical machine learning approaches particularly suitable for high-dimensional genomic datasets with limited sample sizes. Deep learning was therefore explored primarily as a future extension rather than the central modeling strategy of this study.

\section{Handling Small Dataset and Class Imbalance}

Because the TCGA-BRCA cohort contains a limited number of samples relative to the dimensionality of the molecular feature space, several methodological safeguards were incorporated throughout the experimental pipeline to improve model reliability, reduce overfitting risk, and ensure fair evaluation across both ER classes.

\noindent$\bullet$\textbf{Stratified Splitting:} 
All train-test partitions were generated using stratified sampling to preserve the original distribution of ER-positive and ER-negative samples. Maintaining consistent class proportions across training and testing subsets helped prevent minority-class underrepresentation and ensured that model evaluation remained representative of the overall dataset.

\noindent$\bullet$\textbf{Stratified K-Fold Cross-Validation:} To obtain robust and stable performance estimates, all experiments employed five-fold stratified cross-validation. Rather than relying on a single train-test split, the models were evaluated across multiple independent folds while preserving class distributions in each partition. This strategy reduces the variance associated with small datasets and provides a more reliable estimate of model generalization performance. Consequently, the mean performance across all folds was considered a more informative measure than results obtained from a single evaluation split.

\noindent$\bullet$\textbf{Class Weighting:} Given the imbalance between ER-positive and ER-negative samples, most classifiers incorporated balanced class-weighting schemes during training. These weighting strategies increase the importance of minority-class samples and reduce bias toward the majority class. For XGBoost, class imbalance was addressed using:

\begin{equation}
\mathrm{scale\_pos\_weight}
=
\frac{n_{\mathrm{negative}}}{n_{\mathrm{positive}}}
\end{equation}

which assigns additional weight to minority-class observations during optimization and encourages more balanced decision boundaries.

\noindent$\bullet$\textbf{Conservative Hyperparameter Design:} Model hyperparameters were intentionally selected to favor generalization rather than maximize training performance. Regularization mechanisms such as shallow tree depths, feature and sample subsampling, minimum child constraints, conservative learning rates, and penalty terms were incorporated where appropriate. These design choices helped limit model complexity and reduce the likelihood of fitting noise within the high-dimensional feature space. The primary objective was therefore not achieving the highest possible training accuracy, but developing models capable of maintaining stable performance on previously unseen data.

\section{Evaluation Metrics}

Because the TCGA-BRCA dataset exhibits a noticeable class imbalance between ER-positive and ER-negative samples, model performance was evaluated using multiple complementary metrics rather than relying solely on overall accuracy. While standard accuracy provides a general measure of prediction performance, it can be misleading in imbalanced classification settings where strong performance on the majority class may mask poor performance on the minority class. Consequently, this study emphasized metrics that provide a more balanced assessment of classifier behavior across both classes.

\paragraph{\textbf{Balanced Accuracy}} Balanced Accuracy was selected as the primary evaluation metric because it accounts for performance on both positive and negative classes equally. It is computed as:

\begin{equation}
\text{Balanced Accuracy} =
\frac{\text{Sensitivity} + \text{Specificity}}{2}
\end{equation}

Unlike conventional accuracy, Balanced Accuracy is less sensitive to class imbalance and provides a more representative measure of model effectiveness across the entire dataset. For example, a naive classifier that predicts every sample as ER-positive could still achieve relatively high standard accuracy because ER-positive samples dominate the cohort. However, such a model would perform poorly in terms of Balanced Accuracy due to its inability to correctly identify ER-negative patients.

\paragraph{\textbf{ROC-AUC}} The Receiver Operating Characteristic Area Under the Curve (ROC-AUC) was used to evaluate the ability of a model to distinguish between ER-positive and ER-negative samples across a range of classification thresholds. Rather than assessing performance at a single decision boundary, ROC-AUC measures how effectively a model ranks positive samples above negative samples. Higher ROC-AUC values therefore indicate stronger discriminative capability and provide a threshold-independent assessment of model quality.

\paragraph{\textbf{Macro F1-Score}} To further account for class imbalance, Macro F1-score was included as an additional evaluation metric. Unlike weighted averages, the macro formulation assigns equal importance to each class regardless of the number of samples it contains. The F1-score represents the harmonic mean of precision and recall:

\begin{equation}
F1 =
\frac{2 \times \text{Precision} \times \text{Recall}}
{\text{Precision} + \text{Recall}}
\end{equation}

As a result, Macro F1-score provides a balanced measure of classification performance and more strongly penalizes poor performance on minority-class samples.

\paragraph{\textbf{Precision and Recall}} Precision and recall were also reported to provide a more detailed understanding of model behavior. Precision measures the proportion of positive predictions that are correct, thereby reflecting the reliability of positive classifications. Recall, in contrast, measures the proportion of true positive cases that are successfully identified by the model. In medical prediction tasks, recall is particularly important because false-negative predictions may result in patients being incorrectly classified and potentially receiving suboptimal treatment decisions. Reporting both metrics therefore provides additional insight into the trade-off between prediction reliability and case detection.

\section{Hyperparameter Configuration}

Hyperparameters were selected conservatively to reduce overfitting risk under small-data conditions. Given the high-dimensional nature of the molecular feature space and the relatively limited number of patient samples, the primary objective was to promote model generalization rather than maximize training accuracy. Regularization mechanisms, feature subsampling strategies, class-balancing techniques, and conservative optimization settings were incorporated where appropriate to improve robustness across cross-validation folds.

\noindent$\bullet$\textbf{Random Forest:} Random Forest was configured using 300 decision trees together with square-root feature sampling and balanced class weighting. Additional constraints on tree complexity were applied through controlled depth and minimum leaf requirements to reduce overfitting. These settings helped maintain ensemble diversity while limiting excessive memorization of high-dimensional genomic features.

\noindent$\bullet$\textbf{XGBoost:} XGBoost was regularized using shallow tree structures, conservative learning rates, feature subsampling, row subsampling, and minimum child constraints. Class imbalance was handled through the \texttt{scale\_pos\_weight} parameter. Collectively, these design choices improved optimization stability and reduced sensitivity to noisy or redundant molecular variables.

\noindent$\bullet$\textbf{LightGBM and CatBoost:} LightGBM and CatBoost were configured using similar regularization principles. Tree growth was intentionally restricted through constrained depth and limited leaf counts, while subsampling and regularization mechanisms were used to control model complexity. These settings encouraged improved generalization performance and reduced the likelihood of overfitting within the high-dimensional feature space.

\begin{table}[ht]
\centering
\caption{Hyperparameter Configuration}
\label{tab:hyperparameters}
\scriptsize
\renewcommand{\arraystretch}{1.1}
\begin{tabular}{|p{2.3cm}|p{5.2cm}|}
\hline
\textbf{Model} & \textbf{Key Hyperparameters} \\
\hline

Random Forest &
\texttt{n\_estimators=300},
\texttt{max\_features='sqrt'},
\texttt{class\_weight='balanced'}
\\
\hline

XGBoost &
\texttt{max\_depth=shallow},
\texttt{learning\_rate=low},
\texttt{subsample<1.0},
\texttt{scale\_pos\_weight}
\\
\hline

LightGBM &
\texttt{limited num\_leaves},
\texttt{subsample<1.0},
\texttt{L2 regularization}
\\
\hline

CatBoost &
\texttt{depth=constrained},
\texttt{learning\_rate=conservative},
\texttt{verbose=0}
\\
\hline

SVM (RBF) &
\texttt{kernel='rbf'},
\texttt{class\_weight='balanced'},
StandardScaler applied
\\
\hline

Logistic Regression &
\texttt{class\_weight='balanced'},
\texttt{max\_iter=increased},
StandardScaler applied
\\
\hline

Feature Selection &
\texttt{SelectKBest(f\_classif)},
top 300 features selected
\\
\hline

Cross-Validation &
\texttt{StratifiedKFold(n\_splits=5, shuffle=True)}
\\
\hline

\end{tabular}
\end{table}
\noindent$\bullet$\textbf{Support Vector Machine:} The Support Vector Machine employed a radial basis function (RBF) kernel to model nonlinear relationships among molecular features. Prior to training, feature standardization was applied using \texttt{StandardScaler} because kernel-based methods are sensitive to differences in feature magnitude. The nonlinear kernel formulation was particularly useful for CNV experiments, where structural genomic alterations may not follow simple linear relationships.

\noindent$\bullet$\textbf{Logistic Regression:} Logistic Regression served as a linear baseline model and was trained using balanced class weighting together with feature standardization. Although the model offers strong interpretability and computational efficiency, its linear decision boundaries were expected to be less capable of capturing the complex biological interactions associated with ER status. Nevertheless, it provided an important reference point for assessing the benefits of more sophisticated nonlinear learning approaches.

\section{Feature Selection Stability and Biological Interpretation}

Beyond predictive performance, an important objective of this study was to determine whether the selected molecular features reflected meaningful biological signals rather than random statistical artifacts. To evaluate feature stability, the features selected within each cross-validation fold were examined for consistency across repeated experiments. Several genes were repeatedly identified ~\cite{perou2000,sorlie2001} in the RNA-based analyses, including \textit{ESR1}, \textit{PGR}, \textit{FOXA1}, \textit{GATA3}, \textit{TFF1}, and \textit{SCUBE2}. These genes are widely recognized markers of luminal and ER-positive breast cancer subtypes and play central roles in estrogen signaling, hormone responsiveness, and transcriptional regulation.

In contrast, genes such as \textit{KRT5}, \textit{KRT17}, \textit{SOX10}, \textit{GABRP}, and \textit{TRIM29} were consistently associated with ER-negative and basal-like breast cancer phenotypes. The recovery of both ER-positive and ER-negative biological markers across independent cross-validation folds indicates that the feature selection process captured biologically meaningful transcriptional programs rather than dataset-specific noise. This observation is particularly important because feature stability is often challenging in high-dimensional genomic datasets with limited sample sizes.

The repeated identification of well-established breast cancer biomarkers therefore provides additional confidence that the models learned clinically relevant molecular patterns associated with ER status. Consequently, the consistency of feature selection across folds strengthens the robustness, biological validity, and interpretability of the proposed machine learning framework, supporting the reliability of the reported predictive performance.

\section{SINGLE-OMIC EXPERIMENT RESULTS}

\subsection{RNA Expression Results}

RNA expression achieved the strongest predictive performance ~\cite{yoon2020} among all individual omics modalities, consistent with the biological role of Estrogen Receptor (ER) as a transcription factor that directly regulates gene expression. Consequently, transcriptomic profiles provide the most direct molecular representation of ER status.

\begin{table*}[ht]
\centering
\caption{Performance Comparison of Machine Learning Models for ER Status Prediction (RNA-Only)}
\label{tab:model_comparison}
\resizebox{\textwidth}{!}{
\begin{tabular}{lcccccc}
\hline
\textbf{Model} & \textbf{Accuracy (\%)} & \textbf{Balanced Accuracy (\%)} & \textbf{Macro Precision (\%)} & \textbf{Macro Recall (\%)} & \textbf{Macro F1-score (\%)} & \textbf{ROC-AUC (\%)} \\
\hline
Random Forest & $\mathbf{92.35 \pm 1.07}$ & $87.94 \pm 2.18$ & $\mathbf{91.07 \pm 1.82}$ & $87.94 \pm 2.18$ & $\mathbf{89.28 \pm 1.57}$ & $94.60 \pm 2.42$ \\

XGBoost & $91.98 \pm 1.22$ & $\mathbf{88.69 \pm 2.18}$ & $89.60 \pm 1.95$ & $\mathbf{88.69 \pm 2.18}$ & $89.07 \pm 1.67$ & $\mathbf{95.61 \pm 1.50}$ \\

LightGBM & $91.44 \pm 1.79$ & $88.58 \pm 2.03$ & $88.50 \pm 2.90$ & $88.58 \pm 2.03$ & $88.50 \pm 2.32$ & $95.42 \pm 1.97$ \\

CatBoost & $90.89 \pm 2.26$ & $88.22 \pm 2.91$ & $87.58 \pm 3.19$ & $88.22 \pm 2.91$ & $87.85 \pm 2.96$ & $94.80 \pm 1.50$ \\

SVM (RBF) & $91.62 \pm 1.67$ & $88.20 \pm 2.57$ & $89.12 \pm 2.56$ & $88.20 \pm 2.57$ & $88.58 \pm 2.27$ & $93.59 \pm 3.08$ \\

Logistic Regression & $87.98 \pm 1.76$ & $84.54 \pm 1.18$ & $83.78 \pm 2.85$ & $84.54 \pm 1.18$ & $84.06 \pm 1.92$ & $92.38 \pm 1.91$ \\
\hline
\end{tabular}
}
\end{table*}

\begin{table*}[ht]
\centering
\caption{Held-out Test Performance for RNA-Only Models}
\label{tab:rna_results}
\begin{tabular}{lcccccc}
\hline
\textbf{Model} & \textbf{Accuracy} & \textbf{Balanced Accuracy} & \textbf{Macro Precision} & \textbf{Macro Recall} & \textbf{Macro F1-score} & \textbf{ROC-AUC} \\
\hline
Random Forest & \textbf{89.09} & 84.03 & \textbf{86.00} & 84.00 & \textbf{85.00} & 91.34 \\
XGBoost & 87.27 & 84.07 & 83.00 & 84.00 & 83.00 & \textbf{93.49} \\
SVM (RBF) & \textbf{89.09} & 84.03 & \textbf{86.00} & 84.00 & \textbf{85.00} & 88.44 \\
Logistic Regression & 83.64 & 79.16 & 78.00 & 79.00 & 78.00 & 86.17 \\
LightGBM & 88.18 & \textbf{84.67} & 84.00 & \textbf{85.00} & 84.00 & 91.48 \\
CatBoost & 88.18 & \textbf{84.67} & 84.00 & \textbf{85.00} & 84.00 & 93.44 \\
\hline
\end{tabular}
\end{table*}

\begin{figure}[H] 
    \centering 
    \includegraphics[width=1\linewidth]{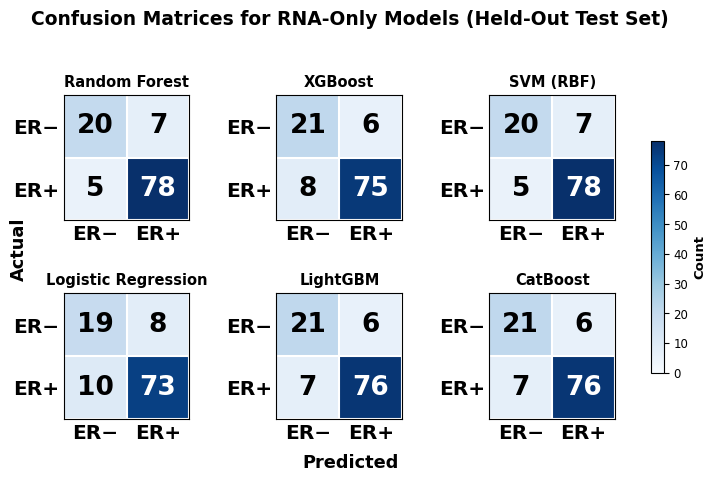} \caption{Confusion Matrix for held-out Set for different ML models for RNA-only Samples} \label{fig:RNA-only-confusion-matrix} 
\end{figure}

Table~\ref{tab:model_comparison} summarizes the five-fold cross-validation performance of all evaluated models. Random Forest achieved the highest classification accuracy ($92.35 \pm 1.07\%$), macro precision ($91.07 \pm 1.82\%$), and macro F1-score ($89.28 \pm 1.57\%$), whereas XGBoost obtained the highest balanced accuracy ($88.69 \pm 2.18\%$) and ROC-AUC ($95.61 \pm 1.50\%$). LightGBM, CatBoost, and SVM demonstrated similarly strong performance, all achieving balanced accuracies above $88\%$, while Logistic Regression consistently produced the weakest results, indicating that nonlinear models better capture the complex transcriptional patterns associated with ER status.

The held-out test results in Table~\ref{tab:rna_results} confirm these trends. Random Forest and SVM achieved the highest test accuracy ($89.09\%$), whereas LightGBM and CatBoost achieved the highest balanced accuracy ($84.67\%$). XGBoost yielded the highest ROC-AUC ($93.49\%$), demonstrating excellent discriminative ability on previously unseen samples. The relatively small reduction from cross-validation to held-out performance indicates good generalization with limited overfitting.

The confusion matrices shown in Fig.~\ref{fig:RNA-only-confusion-matrix} further illustrate model behavior. Ensemble models consistently produced high true-positive rates for ER-positive tumors while maintaining relatively few false positives and false negatives. Random Forest and SVM exhibited nearly identical prediction patterns, whereas XGBoost slightly improved ER-negative detection at the cost of additional false negatives. Logistic Regression produced the largest number of misclassifications, particularly false negatives, explaining its lower balanced accuracy and macro F1-score. Across all models, most errors occurred in the ER-negative class, reflecting the underlying class imbalance despite the use of class-weighted learning and balanced evaluation metrics.

Overall, the RNA-only experiments demonstrate that transcriptomic features provide the most informative molecular representation of ER status. The consistently strong performance of ensemble tree-based models and SVM supports the hypothesis that ER-associated transcriptional programs contain highly discriminative biological signals, making RNA expression the strongest individual modality for ER status prediction in the TCGA-BRCA cohort.

\noindent\textbf{Observation 1:}
RNA expression was the best-performing single-omic modality throughout this study. Random Forest achieved the highest cross-validation accuracy ($92.35 \pm 1.07\%$), while XGBoost achieved the highest ROC-AUC ($95.61 \pm 1.50\%$). On the independent test set, Random Forest and SVM obtained the highest accuracy ($89.09\%$), whereas LightGBM and CatBoost achieved the highest balanced accuracy ($84.67\%$). The confusion matrices further demonstrate consistently high true-positive detection across all models, confirming that transcriptomic features provide the strongest predictive signal for ER status classification.

\subsection{Copy Number Variation Results}

Copy number variation (CNV) provides genomic structural information that is complementary to transcriptomic data for Estrogen Receptor (ER) status prediction. Although CNV alterations are biologically associated with tumor progression, they represent indirect genomic effects and therefore are expected to exhibit lower predictive power than RNA expression.

Table~\ref{tab:model_performance} summarizes the five-fold cross-validation results. Among all evaluated models, SVM with an RBF kernel achieved the best overall performance, obtaining the highest accuracy (91.80 $\pm$ 1.52\%), balanced accuracy (90.20 $\pm$ 2.41\%), macro F1-score (90.20 $\pm$ 2.11\%), and ROC-AUC (95.50 $\pm$ 1.84\%). CatBoost and LightGBM produced competitive balanced accuracies of 82.70\% and 82.50\%, respectively, while Random Forest and XGBoost achieved identical balanced accuracies of 80.40\%. Logistic Regression consistently produced the lowest overall performance, suggesting that linear decision boundaries are insufficient to capture the complex nonlinear relationships present in CNV data.

The held-out test results in Table~\ref{tab:cnv_results} confirm the cross-validation trends. SVM again achieved the strongest generalization performance with a held-out accuracy of 91.82\%, balanced accuracy of 90.20\%, macro F1-score of 90.00\%, and ROC-AUC of 95.50\%. The remaining models achieved balanced accuracies between 80.37\% and 82.67\%, indicating that although CNV contains discriminative information, its predictive signal is weaker than that of RNA expression.

\begin{figure}[!t]
    \centering
    \includegraphics[width=\linewidth]{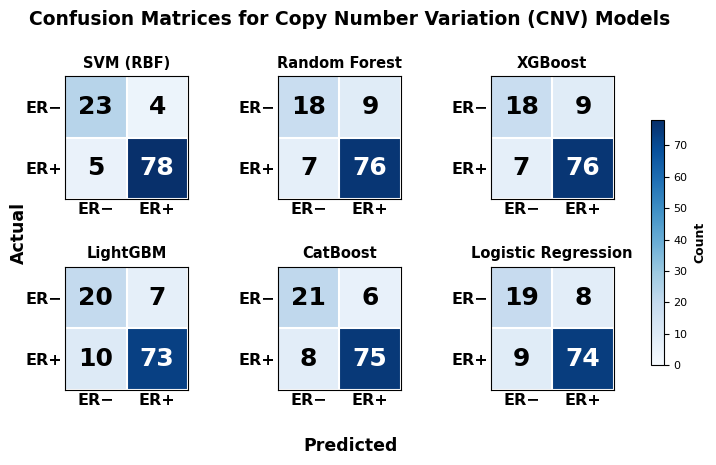}
    \caption{Held-out confusion matrices for the six machine learning models evaluated using CNV features. SVM achieved the best class separation with the fewest false-positive and false-negative predictions, consistent with its superior balanced accuracy and ROC-AUC.}
    \label{fig:cnv_confusion_matrix}
\end{figure}

Fig.~\ref{fig:cnv_confusion_matrix} further illustrates the classification behavior of each model. SVM produced the most balanced confusion matrix, correctly classifying 23 ER-negative and 78 ER-positive samples while generating only four false positives and five false negatives. In contrast, the remaining models exhibited larger numbers of false-positive and false-negative predictions, particularly Random Forest and XGBoost. These results demonstrate the superior ability of nonlinear kernel learning to model structural genomic alterations associated with ER status.

\begin{table*}[ht]
\centering
\caption{Performance Comparison of Machine Learning Models for ER Status Prediction (CNV-Only)}
\label{tab:model_performance}
\resizebox{\textwidth}{!}{
\begin{tabular}{lcccccc}
\hline
\textbf{Model} & \textbf{Accuracy (\%)} & \textbf{Balanced Accuracy (\%)} & \textbf{Macro Precision (\%)} & \textbf{Macro Recall (\%)} & \textbf{Macro F1-score (\%)} & \textbf{ROC-AUC (\%)} \\
\hline
SVM (RBF) & $\mathbf{91.80 \pm 1.52}$ & $\mathbf{90.20 \pm 2.41}$ & $\mathbf{90.10 \pm 2.18}$ & $\mathbf{90.20 \pm 2.41}$ & $\mathbf{90.20 \pm 2.11}$ & $\mathbf{95.50 \pm 1.84}$ \\
Random Forest & $85.50 \pm 2.64$ & $80.40 \pm 3.12$ & $80.60 \pm 3.04$ & $80.40 \pm 3.12$ & $80.40 \pm 2.95$ & $88.20 \pm 2.76$ \\
XGBoost & $85.50 \pm 2.48$ & $80.40 \pm 2.88$ & $80.40 \pm 2.67$ & $80.40 \pm 2.88$ & $80.40 \pm 2.73$ & $88.20 \pm 2.41$ \\
LightGBM & $84.50 \pm 3.02$ & $82.50 \pm 2.57$ & $80.30 \pm 3.11$ & $82.50 \pm 2.57$ & $80.30 \pm 2.96$ & $87.80 \pm 2.64$ \\
CatBoost & $85.50 \pm 2.21$ & $82.70 \pm 2.34$ & $82.30 \pm 2.47$ & $82.70 \pm 2.34$ & $82.30 \pm 2.25$ & $88.30 \pm 2.18$ \\
Logistic Regression & $84.50 \pm 2.88$ & $81.20 \pm 2.96$ & $79.40 \pm 3.25$ & $81.20 \pm 2.96$ & $79.40 \pm 3.01$ & $86.50 \pm 2.71$ \\
\hline
\end{tabular}
}
\end{table*}

\begin{table*}[ht]
\centering
\caption{Held-out Test Performance for Copy Number (CNV) Models}
\label{tab:cnv_results}
\begin{tabular}{lcccccc}
\hline
\textbf{Model} & \textbf{Accuracy} & \textbf{Balanced Accuracy} & \textbf{Macro Precision} & \textbf{Macro Recall} & \textbf{Macro F1-score} & \textbf{ROC-AUC} \\
\hline
SVM (RBF) & \textbf{91.82} & \textbf{90.20} & \textbf{90.00} & \textbf{90.00} & \textbf{90.00} & \textbf{95.50} \\
Random Forest & 85.45 & 80.37 & 80.00 & 80.00 & 80.00 & 88.20 \\
XGBoost & 85.45 & 80.37 & 80.00 & 80.00 & 80.00 & 88.20 \\
LightGBM & 84.55 & 82.45 & 80.00 & 82.00 & 80.00 & 87.80 \\
CatBoost & 85.45 & 82.67 & 82.00 & 82.00 & 82.00 & 88.30 \\
Logistic Regression & 84.55 & 81.20 & 79.00 & 81.00 & 79.00 & 86.50 \\
\hline
\end{tabular}
\end{table*}

\noindent\textbf{Observation 2:}
CNV features provided meaningful but weaker predictive performance than RNA expression. SVM consistently achieved the best cross-validation and held-out performance, demonstrating that nonlinear kernel methods are well suited for modeling structural genomic alterations. The confusion matrices further confirm superior class separation by SVM. Overall, these findings suggest that CNV contributes complementary genomic information for ER status prediction but is more effective as part of a multi-omic framework than as a standalone modality.

\subsection{Protein Expression Results}

Protein expression experiments were performed using RPPA features to evaluate whether downstream proteomic activity contributes to Estrogen Receptor (ER) status prediction. Unlike RNA expression, which directly reflects transcriptional regulation, RPPA measurements capture protein-level pathway activity and functional molecular states. Therefore, protein expression was expected to provide complementary predictive information beyond RNA expression and CNV features.

Table~\ref{tab:rppa_cv_results} summarizes the five-fold cross-validation results. Among all evaluated models, XGBoost achieved the strongest overall RPPA-only performance, obtaining the highest accuracy (91.82 $\pm$ 1.67\%), balanced accuracy (90.20 $\pm$ 2.36\%), macro recall (90.20 $\pm$ 2.36\%), and macro F1-score (89.50 $\pm$ 2.18\%). Random Forest and SVM also demonstrated strong discriminative ability, each achieving a ROC-AUC of 93.10\%, while CatBoost produced competitive balanced performance. Logistic Regression showed the weakest results, indicating that linear decision boundaries were less effective for modeling protein-level molecular patterns associated with ER status.

The held-out test results in Table~\ref{tab:rppa_test_results} confirm the cross-validation trends. XGBoost again achieved the best overall performance, with an accuracy of 91.82\%, balanced accuracy of 90.20\%, macro F1-score of 89.50\%, and ROC-AUC of 93.10\%. CatBoost followed closely with an accuracy of 90.91\% and balanced accuracy of 88.70\%, while Random Forest and SVM maintained stable performance with ROC-AUC values of 93.10\%. These results show that RPPA features contain meaningful discriminative information for ER status classification.

\begin{figure}[!t]
    \centering
    \includegraphics[width=\linewidth]{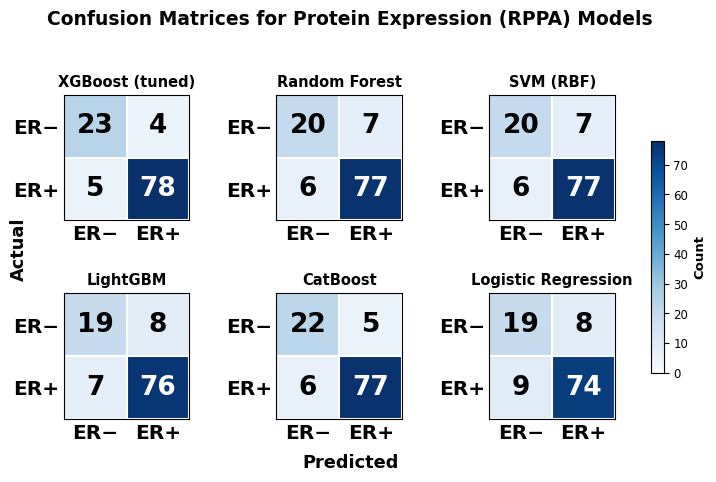}
    \caption{Held-out confusion matrices for the six machine learning models evaluated using RPPA protein expression features. XGBoost achieved the most balanced classification pattern, with the fewest combined false-positive and false-negative predictions.}
    \label{fig:rppa_confusion_matrix}
\end{figure}

Fig.~\ref{fig:rppa_confusion_matrix} further illustrates model behavior on the held-out test set. XGBoost produced the most balanced confusion matrix, correctly classifying 23 ER-negative and 78 ER-positive samples while generating only four false positives and five false negatives. CatBoost also performed competitively, correctly identifying 22 ER-negative and 77 ER-positive samples. In contrast, Logistic Regression produced the largest number of errors, consistent with its lower balanced accuracy and macro F1-score.

\begin{table*}[ht]
\centering
\caption{Five-Fold Cross-Validation Results for Protein Expression (RPPA) ER Status Prediction}
\label{tab:rppa_cv_results}
\resizebox{\textwidth}{!}{
\begin{tabular}{lcccccc}
\hline
\textbf{Model} & \textbf{Accuracy} & \textbf{Balanced Accuracy} & \textbf{Macro Precision} & \textbf{Macro Recall} & \textbf{Macro F1-score} & \textbf{ROC-AUC} \\
\hline
XGBoost (tuned) & $\mathbf{91.82 \pm 1.67}$ & $\mathbf{90.20 \pm 2.36}$ & $\mathbf{89.50 \pm 2.12}$ & $\mathbf{90.20 \pm 2.36}$ & $\mathbf{89.50 \pm 2.18}$ & $\mathbf{93.10 \pm 1.84}$ \\
Random Forest & $90.00 \pm 2.11$ & $85.90 \pm 2.78$ & $86.30 \pm 2.54$ & $85.90 \pm 2.78$ & $86.30 \pm 2.41$ & $\mathbf{93.10 \pm 1.95}$ \\
SVM (RBF) & $90.00 \pm 2.35$ & $85.90 \pm 2.57$ & $86.20 \pm 2.72$ & $85.90 \pm 2.57$ & $86.20 \pm 2.48$ & $\mathbf{93.10 \pm 2.04}$ \\
LightGBM & $88.18 \pm 2.64$ & $84.50 \pm 2.88$ & $83.70 \pm 3.01$ & $84.50 \pm 2.88$ & $83.70 \pm 2.94$ & $91.50 \pm 2.31$ \\
CatBoost & $90.91 \pm 1.88$ & $88.70 \pm 2.22$ & $88.70 \pm 2.11$ & $88.70 \pm 2.22$ & $88.70 \pm 2.09$ & $91.90 \pm 1.74$ \\
Logistic Regression & $84.55 \pm 2.91$ & $82.30 \pm 2.74$ & $80.50 \pm 3.04$ & $82.30 \pm 2.74$ & $80.50 \pm 2.85$ & $87.90 \pm 2.68$ \\
\hline
\end{tabular}
}
\end{table*}

\begin{table*}[ht]
\centering
\caption{Held-out Test Performance for Protein Expression (RPPA) Models}
\label{tab:rppa_test_results}
\resizebox{\textwidth}{!}{
\begin{tabular}{lcccccc}
\hline
\textbf{Model} & \textbf{Accuracy} & \textbf{Balanced Accuracy} & \textbf{Macro Precision} & \textbf{Macro Recall} & \textbf{Macro F1-score} & \textbf{ROC-AUC} \\
\hline
XGBoost (tuned) & \textbf{91.82} & \textbf{90.20} & \textbf{89.50} & \textbf{90.00} & \textbf{89.50} & \textbf{93.10} \\
Random Forest & 90.00 & 85.90 & 86.00 & 86.00 & 86.00 & \textbf{93.10} \\
SVM (RBF) & 90.00 & 85.90 & 86.00 & 86.00 & 86.00 & \textbf{93.10} \\
LightGBM & 88.18 & 84.50 & 83.00 & 84.00 & 83.00 & 91.50 \\
CatBoost & 90.91 & 88.70 & 88.00 & 88.00 & 88.00 & 91.90 \\
Logistic Regression & 84.55 & 82.30 & 80.00 & 82.00 & 80.00 & 87.90 \\
\hline
\end{tabular}
}
\end{table*}

\noindent\textbf{Observation 3:}
RPPA-based protein expression provided strong secondary predictive capability for ER status classification. XGBoost achieved the best overall protein-only performance, with a held-out accuracy of 91.82\%, balanced accuracy of 90.20\%, macro F1-score of 89.50\%, and ROC-AUC of 93.10\%. The confusion matrix further confirmed that XGBoost produced the most balanced classification behavior, correctly identifying 23 ER-negative and 78 ER-positive samples with only four false positives and five false negatives. Overall, RPPA features capture meaningful downstream pathway activity and serve as an important complementary modality, although RNA expression remains the strongest standalone predictor because ER biology is primarily reflected through transcriptional regulation.

\section{Multi-Omic Integration Results}

The multi-omic integration experiment evaluated whether combining transcriptomic (RNA), genomic (CNV), and proteomic (RPPA) information improves Estrogen Receptor (ER) status prediction over individual omics modalities. By integrating complementary molecular features into a unified representation, the models were able to exploit information from multiple biological layers while capturing interactions that are not available from a single modality.

Table~\ref{tab:multiomic_cv_results} summarizes the five-fold cross-validation results. Random Forest achieved the strongest overall performance, obtaining the highest accuracy (93.64 $\pm$ 1.12\%), balanced accuracy (90.30 $\pm$ 2.07\%), macro F1-score (90.20 $\pm$ 1.88\%), and ROC-AUC (97.10 $\pm$ 1.41\%). XGBoost, CatBoost, and SVM also demonstrated competitive performance with identical balanced accuracies of 89.60\%, while LightGBM and Logistic Regression produced slightly lower but still robust results. These findings indicate that nonlinear ensemble methods are particularly effective for integrating heterogeneous molecular features.

The held-out test results in Table~\ref{tab:multiomic_test_results} confirm the cross-validation trends. Random Forest again achieved the best generalization performance with an accuracy of 93.64\%, balanced accuracy of 90.30\%, macro F1-score of 90.00\%, and ROC-AUC of 97.10\%. XGBoost, CatBoost, and SVM each achieved an accuracy of 91.82\% with balanced accuracies of 89.60\%, demonstrating stable performance across unseen samples. Overall, multi-omic integration yielded the strongest predictive performance ~\cite{wang2023,sharma2024} among all experiments conducted in this study.

\begin{figure}[!t]
    \centering
    \includegraphics[width=\linewidth]{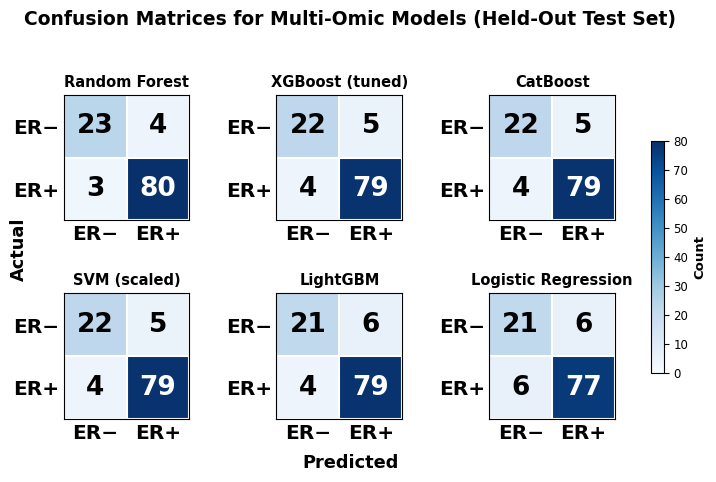}
    \caption{Held-out confusion matrices for the six classical machine learning models evaluated using integrated RNA, CNV, and RPPA features. Random Forest achieved the strongest class separation with the fewest false-positive and false-negative predictions, consistent with its superior balanced accuracy and ROC-AUC.}
    \label{fig:multiomic_cm}
\end{figure}

Fig.~\ref{fig:multiomic_cm} further illustrates the classification behavior of the classical machine learning models. Random Forest produced the most balanced confusion matrix, correctly classifying 23 ER-negative and 80 ER-positive samples while generating only four false positives and three false negatives. XGBoost, CatBoost, and SVM showed highly similar prediction patterns with only minor increases in classification errors, whereas Logistic Regression produced the weakest class separation. These results demonstrate that ensemble-based nonlinear models are better suited for learning complementary information from integrated multi-omic data.

\begin{figure}[!t]
    \centering
    \includegraphics[width=\linewidth]{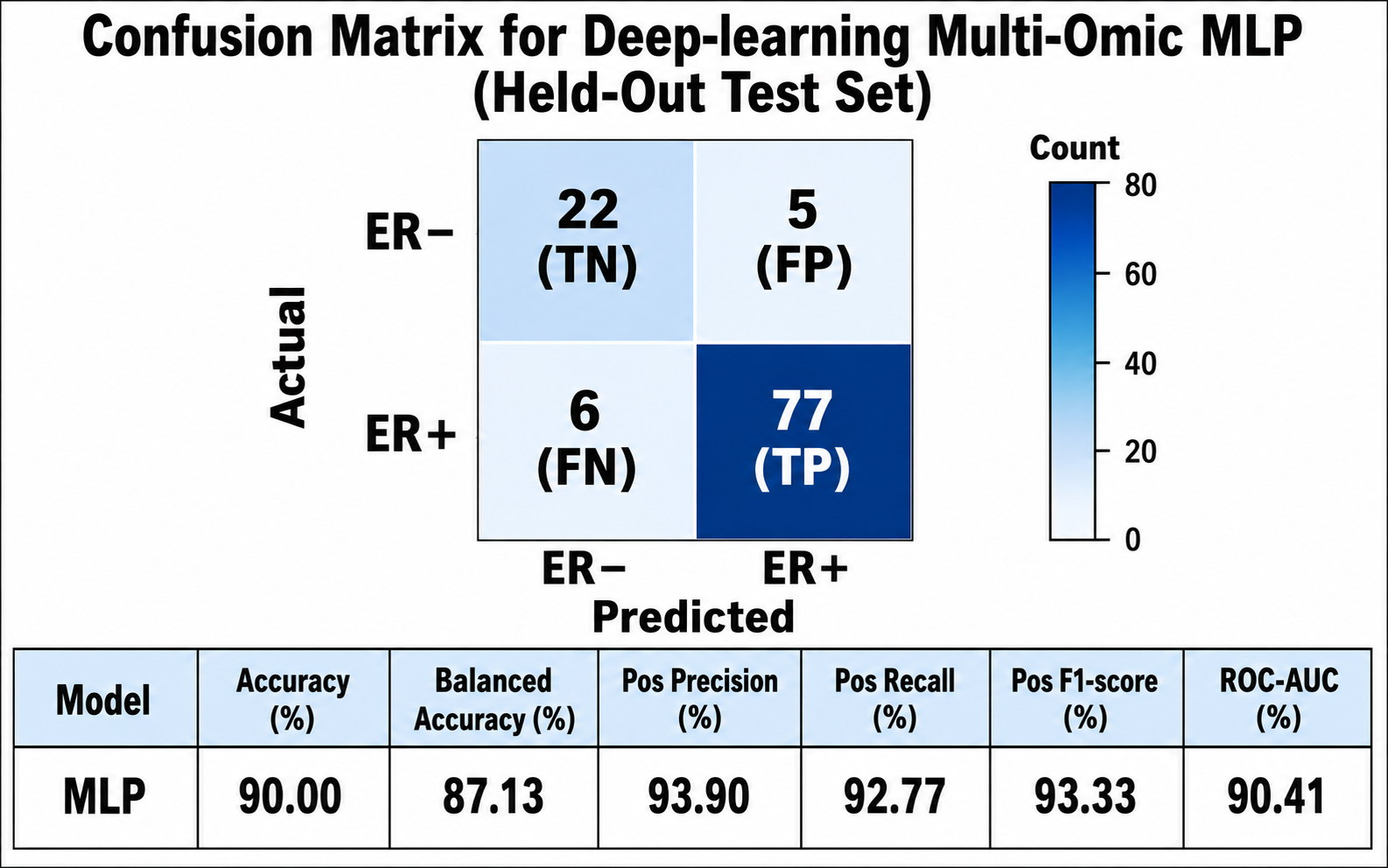}
    \caption{Held-Out confusion matrix for the MLP baseline trained on the integrated multi-omic feature space. Although the MLP achieved competitive performance, it did not surpass the best classical machine learning models.}
    \label{fig:multiomic_mlp_cm}
\end{figure}

To provide a deep learning baseline, a Multi-Layer Perceptron (MLP) was also evaluated using the same integrated feature space. As shown in Fig.~\ref{fig:multiomic_mlp_cm}, the MLP correctly classified 22 ER-negative and 77 ER-positive samples, achieving an accuracy of 90.00\%, balanced accuracy of 87.13\%, macro F1-score of 93.33\%, and ROC-AUC of 90.41\%. Although the MLP produced competitive results, it did not outperform the best classical machine learning models. This observation is consistent with the relatively small sample size and high-dimensional feature space, where tree-based ensemble methods generally exhibit better generalization than deep neural networks.

\begin{table*}[ht]
\centering
\caption{Five-Fold Cross-Validation Results for Multi-Omic ER Status Prediction (RNA + CNV + RPPA)}
\label{tab:multiomic_cv_results}
\resizebox{\textwidth}{!}{
\begin{tabular}{lcccccc}
\hline
\textbf{Model} & \textbf{Accuracy} & \textbf{Balanced Accuracy} & \textbf{Macro Precision} & \textbf{Macro Recall} & \textbf{Macro F1-score} & \textbf{ROC-AUC} \\
\hline
Random Forest & $\mathbf{93.64 \pm 1.12}$ & $\mathbf{90.30 \pm 2.07}$ & $\mathbf{90.20 \pm 1.96}$ & $\mathbf{90.30 \pm 2.07}$ & $\mathbf{90.20 \pm 1.88}$ & $\mathbf{97.10 \pm 1.41}$ \\
XGBoost (tuned) & $91.82 \pm 1.56$ & $89.60 \pm 2.31$ & $89.20 \pm 2.14$ & $89.60 \pm 2.31$ & $89.20 \pm 2.05$ & $93.40 \pm 1.92$ \\
CatBoost & $91.82 \pm 1.74$ & $89.60 \pm 2.18$ & $89.20 \pm 2.09$ & $89.60 \pm 2.18$ & $89.20 \pm 2.11$ & $93.60 \pm 1.76$ \\
SVM (scaled) & $91.82 \pm 1.63$ & $89.60 \pm 2.44$ & $89.20 \pm 2.28$ & $89.60 \pm 2.44$ & $89.20 \pm 2.17$ & $90.00 \pm 2.84$ \\
LightGBM & $90.91 \pm 1.95$ & $88.90 \pm 2.36$ & $88.40 \pm 2.52$ & $88.90 \pm 2.36$ & $88.40 \pm 2.27$ & $91.70 \pm 2.08$ \\
Logistic Regression & $89.09 \pm 2.11$ & $87.80 \pm 2.64$ & $86.30 \pm 2.73$ & $87.80 \pm 2.64$ & $86.30 \pm 2.48$ & $91.50 \pm 2.17$ \\
\hline
\end{tabular}
}
\end{table*}

\begin{table*}[ht]
\centering
\caption{Held-out Test Performance for Multi-Omic Models}
\label{tab:multiomic_test_results}
\resizebox{\textwidth}{!}{
\begin{tabular}{lcccccc}
\hline
\textbf{Model} & \textbf{Accuracy} & \textbf{Balanced Accuracy} & \textbf{Macro Precision} & \textbf{Macro Recall} & \textbf{Macro F1-score} & \textbf{ROC-AUC} \\
\hline
Random Forest & \textbf{93.64} & \textbf{90.30} & \textbf{90.00} & \textbf{90.00} & \textbf{90.00} & \textbf{97.10} \\
XGBoost (tuned) & 91.82 & 89.60 & 89.00 & 89.00 & 89.00 & 93.40 \\
CatBoost & 91.82 & 89.60 & 89.00 & 89.00 & 89.00 & 93.60 \\
SVM (scaled) & 91.82 & 89.60 & 89.00 & 89.00 & 89.00 & 90.00 \\
LightGBM & 90.91 & 88.90 & 88.00 & 88.00 & 88.00 & 91.70 \\
Logistic Regression & 89.09 & 87.80 & 86.00 & 87.00 & 86.00 & 91.50 \\
\hline
\end{tabular}
}
\end{table*}

\noindent\textbf{Observation 4:}
Multi-omic integration achieved the strongest predictive performance in this study. Random Forest consistently outperformed all evaluated models, obtaining a held-out accuracy of 93.64\%, balanced accuracy of 90.30\%, macro F1-score of 90.00\%, and ROC-AUC of 97.10\%. The confusion matrix further confirmed its superior classification performance, correctly identifying 23 ER-negative and 80 ER-positive samples with only four false positives and three false negatives. Although the MLP baseline achieved competitive results, classical ensemble methods generalized better on this small, high-dimensional dataset. Overall, integrating RNA, CNV, and RPPA features provided complementary molecular information, yielding modest but consistent improvements over individual omics modalities.

\section{Limitations}

This study has several limitations. First, although the TCGA-BRCA cohort provides comprehensive transcriptomic, genomic, and proteomic information, the final dataset contained only 549 usable patient samples. The combination of a limited sample size and high-dimensional feature space increases the risk of overfitting despite the use of stratified cross-validation, fold-specific feature selection, class weighting, and conservative hyperparameter tuning.

Second, model evaluation was performed using stratified train--test splitting and five-fold cross-validation within the TCGA-BRCA cohort. While these procedures provide reliable internal performance estimates, external validation on independent cohorts such as METABRIC or GEO is necessary to assess generalizability across different patient populations, sequencing platforms, and preprocessing pipelines.

Third, deep learning was explored only through a MLP baseline. More advanced architectures were not extensively investigated because deep neural networks generally require substantially larger training datasets to achieve reliable generalization on high-dimensional genomic data. Consequently, classical machine learning methods were better suited to the current experimental setting.

Finally, this study considered three omics modalities: RNA expression, copy number variation (CNV), and protein expression (RPPA). Additional molecular and clinical information, including DNA methylation, somatic mutations, microRNA expression, pathway-level features, and patient clinical characteristics, may further improve predictive performance and biological interpretation. Future work will focus on integrating these modalities, validating the proposed framework on independent cohorts, and investigating advanced multimodal deep learning approaches for precision oncology.

\section{Conclusion}

This study presented a systematic benchmarking analysis of classical machine learning models for Estrogen Receptor (ER) status prediction using multi-omics data from the TCGA-BRCA cohort. RNA expression, copy number variation (CNV), and protein expression (RPPA) were evaluated individually and through multi-omic integration using a rigorous experimental framework incorporating stratified data splitting, fold-specific feature selection, class imbalance handling, and balanced performance evaluation.

Among the individual modalities, RNA expression consistently provided the strongest predictive performance, reflecting the central role of ER as a transcription factor. CNV and RPPA contributed complementary genomic and proteomic information that further improved prediction when integrated with transcriptomic features. Overall, multi-omic integration achieved the best predictive performance, demonstrating that combining complementary molecular information enhances model robustness while providing modest but consistent improvements over single-omic models.

Across all evaluated algorithms, nonlinear machine learning methods consistently outperformed the linear baseline. Random Forest achieved the strongest overall performance, obtaining a balanced accuracy of 90.30\%, ROC-AUC of 97.10\%, and the best confusion matrix performance on the held-out test set. In comparison, the MLP baseline achieved competitive but lower performance, indicating that carefully regularized classical machine learning methods remain highly effective for small, high-dimensional genomic datasets.

The repeated selection of well-established ER-associated biomarkers, including \textit{ESR1}, \textit{PGR}, \textit{FOXA1}, and \textit{GATA3}, further supports the biological validity and interpretability of the proposed framework. These findings demonstrate that robust machine learning models can achieve strong predictive performance while preserving clinically meaningful biological insights.

Overall, this work provides a comprehensive benchmark for multi-omics-based ER status prediction and demonstrates the effectiveness of integrating transcriptomic, genomic, and proteomic information using classical machine learning. Future work will focus on validating the proposed framework using independent cohorts, incorporating additional omics and clinical modalities, and investigating advanced multimodal deep learning architectures to further improve predictive performance and clinical applicability.

\section{Future Work}

Several directions may further extend this work and improve both predictive performance and biological interpretability. First, future studies should incorporate additional molecular and clinical modalities, including DNA methylation, somatic mutations, microRNA expression, pathway-level features, and patient clinical characteristics. Integrating these complementary data sources may provide a more comprehensive representation of breast cancer biology and further enhance ER status prediction.

Second, external validation using independent breast cancer cohorts, such as METABRIC and GEO, is necessary to evaluate the robustness and generalizability of the proposed framework across different patient populations, sequencing platforms, and preprocessing pipelines. Such validation is an important step toward potential clinical translation.

Another promising direction is the incorporation of biological prior knowledge into machine learning models. Graph neural networks ~\cite{mogcn2022}, pathway-informed learning, and protein--protein interaction networks may enable models to better capture the complex biological relationships among genes and proteins while improving interpretability and reducing reliance on dataset-specific statistical patterns.

Future work may also extend the prediction task beyond binary ER status classification to clinically relevant molecular subtype prediction, including Luminal A, Luminal B, HER2-enriched, and Basal-like breast cancer. Such multi-class prediction frameworks could provide more detailed molecular characterization and better support precision oncology applications.

Finally, as larger multi-omics datasets become available, advanced multimodal deep learning architectures, including attention-based networks, graph-based models, and late-fusion strategies, should be investigated. These approaches have the potential to better exploit complementary information across multiple omics modalities while preserving modality-specific representations.

Overall, these research directions may improve predictive accuracy, model robustness, biological interpretability, and clinical applicability, contributing to the development of more reliable computational frameworks for multi-omics precision medicine in breast cancer.

\end{document}